\documentclass[runningheads]{llncs}
\usepackage{amssymb}
\usepackage{graphicx}
\usepackage{amsmath}
\usepackage{commath}
\usepackage{float}
\usepackage{blindtext}
\usepackage{scrextend}
\usepackage{multirow}
\begin{document}
\title{Are Registration Uncertainty and Error Monotonically Associated?}
%

%

\author{Jie Luo\inst{1,2}
	\and Sarah Frisken\inst{1}
	\and Duo Wang \inst{1}
	\and Alexandra Golby\inst{1} 
	\and Masashi Sugiyama\inst{5,2} 
	\and William Wells III\inst{1} 
}

\authorrunning{j. Luo et al.}

%


\institute{
	Brigham and Women's Hospital, Harvard Medical School, USA 
	\and Graduate School of Frontier Sciences, The University of Tokyo, Japan 
	\and Department of Automation, Tsinghua University, China 
	\and Center for Advanced Intelligence Project, RIKEN, Japan\\
	\email{jluo5@bwh.harvard.edu}
}

\maketitle

\begin{abstract}
In image-guided neurosurgery, current commercial systems usually provide only rigid registration, partly because it is harder to predict, validate and understand non-rigid registration error. For instance, when surgeons see a discrepancy in aligned image features, they may not be able to distinguish between registration error and actual tissue deformation caused by tumor resection. In this case, the spatial distribution of registration error could help them make more informed decisions, e.g., ignoring the registration where the estimated error is high. However, error estimates are difficult to acquire. Probabilistic image registration (PIR) methods provide measures of registration uncertainty, which could be a surrogate for assessing the registration error. It is intuitive and believed by many clinicians that high uncertainty indicates a large error. However, the monotonic association between uncertainty and error has not been examined in image registration literature. In this pilot study, we attempt to address this fundamental problem by looking at one PIR method, the Gaussian process (GP) registration. We systematically investigate the relation between GP uncertainty and error based on clinical data and show empirically that there is a weak-to-moderate positive monotonic correlation between point-wise GP registration uncertainty and non-rigid registration error.

\end{abstract}
\section{Introduction}
In image-guided neurosurgery (IGN), surgical procedures are often planned based on the preoperative (\textit{p}-) magnetic resonance imaging (MRI). During surgery, clinicians may acquire intraoperative (\textit{i}-) MRI and/or Ultrasound (US). Image registration can be used \cite{Maintz,Sotiras} to map the \textit{p}-MRI to the intraoperative coordinate space to help surgeons locate structures or boundaries of interest during surgery (e.g., tumor margins or nearby blood vessels to be avoided) and facilitate more complete tumor resection \cite{Gerard,Morin,Ma}.

\sloppy
Even though the brain clearly undergoes non-linear deformation during surgery, rigid registration is still the standard for clinical practice \cite{Rivaz}. Although non-rigid registration has long been a goal for IGNs, this goal is hampered because non-rigid registration error is less predictable and harder to validate than rigid registration error. In practice, if surgeons see a discrepancy between two aligned image features, they may not be able to tell if the misalignment is caused by a registration error or an actual tissue deformation caused by tumor resection. In this case, providing surgeons with a spatial distribution of the expected registration error could help them make more informed decisions, e.g., ignoring the registration where the expected error is high. However, determining this spatial distribution is difficult since:

\begin{enumerate}
	\item Most methods that estimate registration error, such as bootstrapping \cite{Kybic,Shams}, perturbed input \cite{Datteri,Hub,Hub2}, stereo confidence \cite{Saygill} and supervised learning \cite{Sokooti,Sokooti2,Saygill2}, require multiple runs of a non-rigid registration algorithm, thus they are too time-consuming to be practical for IGNs where feedback is required within a few minutes of intraoperative image acquisition.
	\item More importantly, existing methods estimate the error by detecting misaligned image features \cite{Shams,Saygill,Sokooti2,Saygill2}. These methods fail in IGNs because tumor resection and retraction significantly alter the brain, particularly at the tumor margin where precision is most needed. Thus finding consistent image features near the tumor margin may be difficult.
\end{enumerate}

\begin{figure*}[t]
	\centering
	\includegraphics[height=3.6cm]{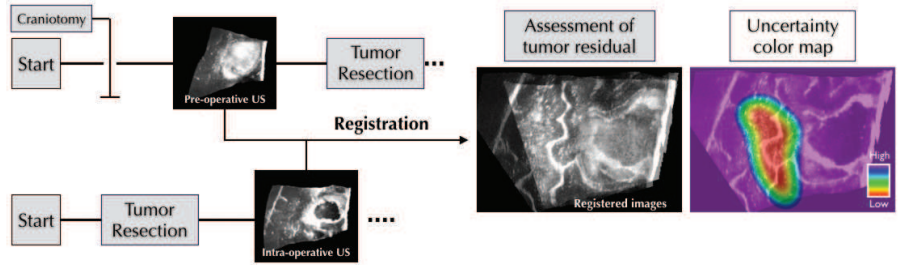}
	\vspace{-3mm}
	\caption{ An example to illustrate the usefulness of registration uncertainty in IGNs.}
	\vspace{-3mm}
\end{figure*}

An alternative for directly estimating the registration error is to use registration uncertainty as a surrogate. Registration uncertainty is a measure of confidence in the predicted registration and typically estimated by probabilistic image registration (PIR)\cite{Glocker,Popuri,Risholm,Lotfi,Wassermann,Yang,Simpson2,Heinrich,Folgoc,Adrian,Miaomiao2,Jax,Sedghi,Siming,Leemput}. In IGN, utilizing registration uncertainty can be helpful. As shown in Fig.1, surgeons can inspect the residual tumor after registering the \textit{p}-US image to the \textit{i}-US image and decide whether to continue the resection or end the operation. An uncertainty color map overlaid on top of the registered images, where red indicated regions of low uncertainty, can be used by surgeons to dismiss clear misregistration regions where uncertainty is high and have more confidence in the registration where uncertainty is low (e.g., red regions).

In this example, surgeons might have higher confidence inside red regions because they assume that areas with low uncertainty also tend to have a low error.  However, this assumption is only valid if the registration uncertainty and error have a positive monotonic association. While this notion is intuitive and believed by many clinicians, to the best of our knowledge, it has not been examined in the PIR literature.

``Are registration uncertainty and error monotonically associated?" is a crucial question that impacts the applicability of registration uncertainty. In this pilot study, we attempt to address this question by looking at a promising PIR method, Gaussian process (GP) registration \cite{Wassermann,Jax,Siming}. We systematically investigate the GP uncertainty and error using point-wise posterior predictive checking and a patch-wise correlation test. We note that the registration uncertainty can be categorized as transformation uncertainty or label uncertainty \cite{Jax2}. Since the applicability of label uncertainty is still in question, this paper will focus solely on the transformation uncertainty when it refers to 'registration uncertainty'.

\section{Methods}

In this section, we briefly review GP registration uncertainty. Then we introduce Spearman's correlation coefficient and provide details about our point-wise and patch-wise experiments.

\subsection{Review of the GP registration uncertainty}

The stochastic GP registration approach has shown promising results in IGNs \cite{Jax,Wassermann,Siming}. As shown in Fig.2, a key step in the GP registration is to estimate $N_*$ unknown displacement vectors $\mathbf{D}_*$ from $N$ known ones $\mathbf{D}$ that were derived from automatic feature extraction and matching.

Let $\mathbf{x}$ be the grid coordinate and $\mathbf{d}(\mathbf{x})=[d_\mathrm{x},d_\mathrm{y},d_\mathrm{z}]$ be the associated displacement vector. For $d(\mathbf{x})$ being one of $d_\mathrm{x}$, $d_\mathrm{y}$, and $d_\mathrm{z}$, it is modeled as a joint Gaussian distribution $d(\mathbf{x})\sim\mathrm{GP}(\mathit{m}(\mathbf{x}),\mathrm{k}(\mathbf{x},\mathbf{x}'))$ with mean function $\mathit{m}(\mathbf{x})=0$ and covariance function  $\mathrm{k}(\mathbf{x},\mathbf{x}')$. Thus $\mathbf{D}$ and $\mathbf{D}_*$ have the following relationship:

\begin{equation}
\left[ \begin{array}{cc} \mathbf{D}\\\mathbf{D_*} \end{array} \right] \sim \mathit{N}\left(\mathbf{m}(\mathbf{x}) , \left[ \begin{array}{ccc} 	\mathbf{K} &	  \mathbf{K}_*\\
\mathbf{K}^T_* &	  \mathbf{K}_{**}.\\ \end{array} \right] \right).
\end{equation}

\begin{figure*}[t]
	\centering
	\includegraphics[height=3.5cm]{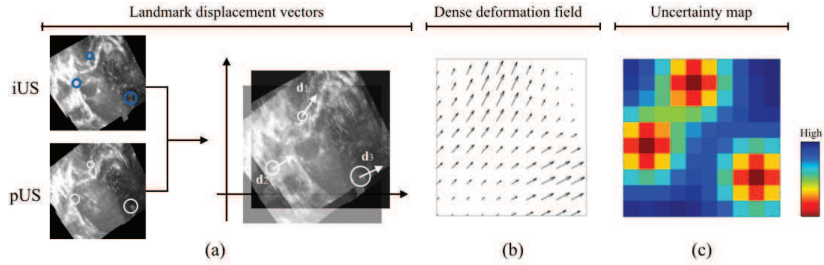}
	\vspace{-3mm}
	\caption{(a) 3 displacement vectors; (b) A $10\times10$ interpolated dense deformation field; (b) A visualization of registration uncertainty.}
	\vspace{-3mm}
\end{figure*}

In Eq.(1), $\mathbf{K}=\mathrm{k}(\mathbf{X},\mathbf{X}) \in \mathbb{R}^{N\times N}$ and $\mathbf{K_{**}}=\mathrm{k}(\mathbf{X_*},\mathbf{X_*}) \in \mathbb{R}^{N_*\times N_*}$ are intra-covariance matrices of $\mathbf{d}$ and $\mathbf{d}_*$ respectively. $\mathbf{K}_*=\mathrm{k}(\mathbf{X},\mathbf{X_*}) \in \mathbb{R}^{N\times N_*}$ is the inter-covariance matrix. The interpolated displacement vector values can be estimated from the mean $\mu_*$ of the posterior distribution of  $p(\mathbf{D}_*\mid\mathbf{X_*},\mathbf{X},\mathbf{D})$:

\begin{equation}
\mu_*= \mathbf{K}^T_*\mathbf{K}^{-1}\mathbf{D}.
\end{equation}

From Eq.(1), the posterior covariance matrix can also be derived as

\begin{equation}
\mathbf{\Sigma}_*= \mathbf{K}_{**}-\mathbf{K}^T_*\mathbf{K}^{-1}\mathbf{K}_*.
\end{equation}

Diagonal entries of $\mathbf{\Sigma_*}$ are the marginal transformation variances, and they can be used as the GP registration uncertainty. In this study, we choose the same kernel $\mathrm{k}(\mathbf{x},\mathbf{x}')=\exp(- \frac{x^2}{a})$ for all three displacement components. 

Fig.2(b) shows a $10\times10$ dense deformation field interpolated from three landmark displacement vectors. Each voxel is associated with an estimated displacement vector and uncertainty value. Fig.2(c) is an uncertainty color map for the displacement field.

\subsection{Spearman's rank correlation coefficient}

Spearman's correlation coefficient, often denoted by $\rho_{\mathrm{s}}$, is a non-parametric measure of statistical dependence between the rankings of two variables. It assesses how well their relationship can be described using a monotonic function \cite{Spearman}.

In this study we prefer $\rho_{\mathrm{s}}$ over Pearson's correlation $\rho_{\mathrm{p}}$ for the following reasons: 

\begin{enumerate}
	\item $\rho_{\mathrm{p}}$ measures the strength of a linear relationship. To be clinically useful, registration uncertainty does not have to be linearly correlated with the error. In this sense, we prefer $\rho_{\mathrm{s}}$ which measures a less ``restrictive" monotonic relationship;
	\item Since $\rho_{\mathrm{s}}$ limits the influence of outliers to the value of its rank, it is less sensitive than $\rho_{\mathrm{p}}$ to strong outliers that lie in the tails of the distribution \cite{Spearman}.
\end{enumerate}

Assume there are $M$ test points, $\mathrm{u}(i)$ and $\epsilon(i)$, which represent the uncertainty and error for point $i$ respectively. Let $U$ and $E$ denote discrete random variables with values $\{\mathrm{u}(1),\mathrm{u}(2), ..., \mathrm{u}(M)\}$ and $\{\epsilon(1),\epsilon(2), ..., \epsilon(M)\}$. To measure $\rho_{\mathrm{s}}$, we have to convert $U$ and $E$ to descending rank vectors $rU$ and $rE$, i.e., the rank vector for $[0.2, 1.2, 0.9, 0.5, 0.1]$ would be $[2, 5, 4, 3, 1]$. Then $\rho_{\mathrm{s}}$ can be estimated as
\begin{equation}
\rho_{\mathrm{s}}= \frac{\mathrm{cov}{(rU,rE)}}{\sigma_{rU}\sigma_{rE}},
\end{equation}
where $\mathrm{cov}$ is the covariance, $\sigma$'s are the standard deviations.  Noticing that $\rho_{\mathrm{s}}$ is by design constrained as $ -1 \leq \rho_{\mathrm{s}} \leq 1$, and 1 indicates a perfect positive monotonic relationship.

\subsection{Point-wise posterior predictive checking}

When a surgeon is removing a tumor mass near a critical structure, it is vital that s/he knows how close the predicted instrument location is from the structure and how confident the prediction is. With GP registration, we can predict the instrument location using a displacement vector. Meanwhile, we can also provide the registration uncertainty to indicate how likely the estimated instrument location is accurate. Here, we designed a point-wise experiment to investigate whether the true location is close to the predicted location when the uncertainty is low and vice versa.

The point-wise experiment is inspired by posterior predictive checking (PPC) \cite{Bayes}. PPC examines the fitness of a model using the similarity between values generated by the posterior distribution and the observed ones.

\begin{figure}[t]
	\centering
	\includegraphics[width=12.0cm]{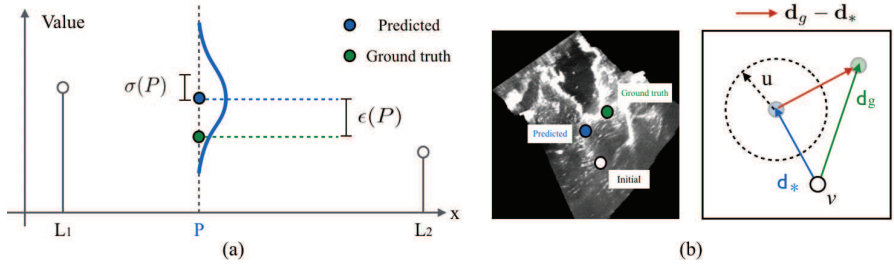}
	\caption{(a) An illustrative example for the point-wise posterior predictive checking experiment; (b) An illustration for how to compute $\epsilon$ and $u$ in the context of IGNs}
	\vspace{-3mm}
\end{figure}

In an illustrative 1D example shown in Fig.3(a), $\mathrm{L}_1$ and $\mathrm{L}_2$ are two landmarks whose values are indicated by the length of vertical bars. The goal is to interpolate the value at location $\mathrm{P}$. Here the blue bell-curve is the estimated posterior distribution $p(\mathrm{P}|\mathrm{L}_1,\mathrm{L}_2)$ and it has a mean of $\mathrm{p}_*$. Since we know the ground truth value $\mathrm{p}_\mathrm{g}$, we can compute the estimation error as $\epsilon(\mathrm{P})=\abs{\mathrm{p}_\mathrm{g}-\mathrm{p}_*}$. The standard deviation $\sigma$ of the posterior is often used to represent the uncertainty $u$ of the estimation.

In the context of GP registration shown in Fig.3(b), the white circle is the initial location of voxel $\mathrm{v}$ on the \textit{p}-US image. The green circle represents the ground truth location of deformed $\mathrm{v}$ on the \textit{i}-US image and the blue circle is the predicted location. In Fig.4(b), $\mathbf{d}_\mathrm{g}$ and $\mathbf{d}_*$ are the ground truth and predicted displacement vectors respectively, and the registration error can be computed as $\epsilon=\norm{\mathbf{d}_\mathrm{g}(i)-\mathbf{d}_*(i)}$. As $\mathrm{u}$ is the registration uncertainty associated with $d_*$, it is visualized by a circle where the magnitude of $\mathit{u}$ is the radius of the circle (a larger circle indicates a higher uncertainty). 

In the point-wise experiment, we compute $u$ and $\epsilon$ for every voxel-of-interest and form two discrete variables $U$ and $E$. Using $\rho_{\mathrm{s}}$, we can measure how strong their monotonic relationship is.

\subsection{Patch-wise correlation test}

We also investigate the correlation between $u$ and $\epsilon$ over image-patches. Because in IGNs, surgeons may be more interested in registration errors over region of interest. We plan to present the uncertainty to surgeons via color overlays so that they can get a higher level understanding of registration error.

Given a voxel $\mathrm{v}$ located at $\mathbf{x}_{\mathrm{v}}\in\mathbb{R}^3$, we define an image patch $\Omega\subset\mathbb{R}^3$ as a sub-volume centered at $\mathbf{x}_{\mathrm{v}}$, let $\Omega$ have size $N$. Assuming $\mathrm{u}_\mathbf{x}$ is the voxel-wise uncertainty at location $\mathbf{x}$, we can compute the patch-wise uncertainty as the mean voxel uncertainty over $\Omega$ as $ \mathit{u}(\Omega)=\frac{1}{\mathrm{N}}\sum_{\mathbf{x}\in\Omega}\mathit{u}_\mathbf{x}$. The estimation of $\epsilon$ over a patch is not straightforward. An ideal way for measuring the patch-wise registration error would be to use the residual Euclidean distance over densely-labeled and well-distributed landmarks placed on both patches. However, to our knowledge, none of the existing neurosurgical datasets has such landmarks.

In this study, since all experiments are based on uni-modal registration, we use  intensity-based dissimilarity metrics to measure the error between ground truth patches $\Omega_\mathrm{g}$'s and predicted patches $\Omega_*$'s. In a previous study that attempted to use patch-wise dissimilarity measures to indicate registration quality, the Histogram Intersection (HI) metric achieved the best result \cite{Vis}. Therefore, we use HI as a dissimilarity metric together with the commonly known Sum of Squared Differences (SSD).

For $\Omega_\mathrm{g}$ and $\Omega_*$, let $\mathit{p}(\mathit{t})$ and $\mathit{q}(\mathit{t})$ be the intensity probability mass functions. $K$ is the number of intensity bins in the histogram. HI can be estimated as
\begin{equation}
HI(\Omega_*,\Omega_\mathrm{g})=1-\sum_{i=1}^{K}\min(p(\mathit{t}_i),q(\mathit{t}_i)).
\end{equation}

\begin{figure}[t]
	\centering
	\includegraphics[width=11.0cm]{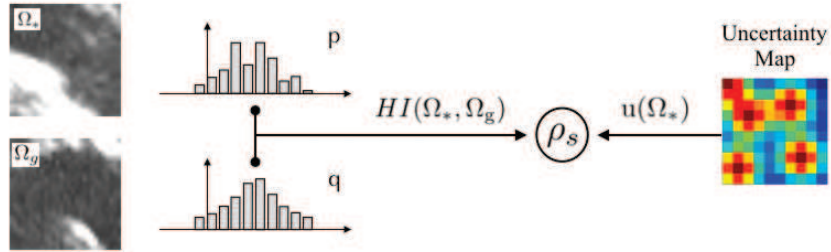}
	\vspace{-3mm}
	\caption{An illustration for using the HI metric to compute Spearman's rank correlation coefficient for patches.}
	\vspace{-2mm}
\end{figure}

Fig.4 illustrates using HI in the patch-wise correlation test. For SSD and HI, their scalar outputs are used as $\epsilon(\Omega)$ for estimating $\rho_{\mathrm{s}}$. Noticing that the size of patches may influence the test result, thus we conduct multiple patch-wise experiments using different patch sizes.

\section{Experiments}
\vspace{-2mm}
We conducted the experiments on two clinical datasets for neurosurgical registration, RESECT \cite{RESECT} and MIBS. RESECT is a public benchmark dataset for IGN \cite{CURIOUS}, while MIBS is a proprietary dataset from a local hospital. Both datasets in total contain 23 sets of \textit{p}-US and \textit{i}-US scans that were acquired from patients with brain tumors. US data were provided as a reconstructed 3D volume. In the \textit{p}-US to \textit{i}-US GP registration context, we tested manually annotated landmarks in the RESECT dataset and automatically detected landmarks in MIBS [35,36], which does not have manual annotations. Noticing that all tested points were not used for GP interpolation. 
\vspace{-2mm}
\subsection{Point-wise experiment}

In the point-wise experiment, for each landmark on the \textit{i}-US image, GP registration estimated $\mathbf{d}_*$ and $\sigma$. Since $\mathbf{d}_\mathrm{g}$ is known, we can calculate $\epsilon$ and then combine all points to compute $\rho_{\mathrm{s}}$ for a pair of images. 

\begin{figure*}[t]
	\centering
	\includegraphics[height=3.5cm]{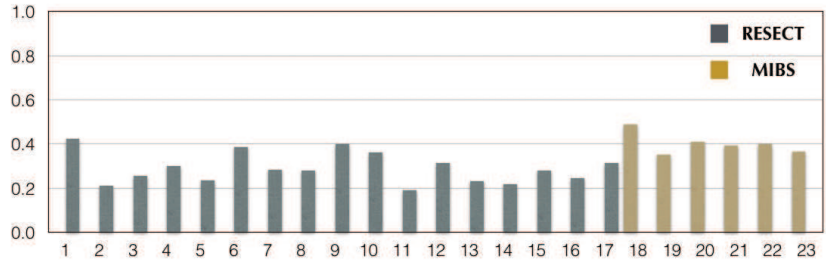}
	\vspace{-3mm}
	\caption{The estimated $\rho_{\mathrm{s}}$ for the point-wise experiment. We can see a moderate positive monotonic relationship between $u$ and $\epsilon$. }
	\vspace{-3mm}
\end{figure*}

The estimated point-wise $\rho_{\mathrm{s}}$'s are summarized in Fig.5.  For manual landmarks in RESECT, the mean value of $\rho_{\mathrm{s}}$ is 0.2899, which indicates a weak-to-moderate positive monotonic correlation. Automatically extracted landmarks achieved an average $\rho_{\mathrm{s}}$ of 0.4014, which can be categorized as a moderate-to-strong correlation. However, both scores are significantly lower than what is required for a perfect positive monotonic relationship. At this stage, it's still too early to conclude definitively whether it is safe to use GP registration uncertainty to assess the accuracy of predicted, e.g., instrument location. 

We suspect that the $\rho_{\mathrm{s}}$ discrepancy between these two groups of landmarks is due to the nature of GP uncertainty and the distribution of landmarks: In GP registration, the uncertainty of a voxel depends on its distance to neighboring interpolating points, e.g., the closer to interpolating points, the lower uncertainty it has. If an annotated landmark is far away from all interpolating landmarks, it is likely to have high uncertainty. In case it happens to be located in a region with less severe deformation, that highly uncertain landmark would have a low registration error, thus lower the overall score for $\rho_{\mathrm{s}}$.

\vspace{-2mm}
\subsection{Patch-wise experiment}

In the patch-wise experiment, we padded $\pm k$ surrounding voxels to each landmark. For example, $\pm2$ padding generates a patch of the size of $5\times5\times5$. Tested values of $k$ include 3 and 5. We calculated $\epsilon$ using SSD/HI for all patches and computed $\rho_{\mathrm{s}}(\Omega)$ afterward. 

The estimated patch-wise $\rho_{\mathrm{s}}$'s are shown in Fig.6. It can be seen that values of $\rho_{\mathrm{s}}(\Omega)$'s are consistently low for both datasets.  We deduce the reasons for low $\rho_{\mathrm{s}}(\Omega)$ values are: (1) In the presence of large deformation, e.g., tumor resection, a pair of well-matched patches may look drastically different. In this case, instead of the residual Euclidean distance over densely-labeled and well-distributed landmarks, other appearance-based dissimilarity measures become sub-optimal for estimating the registration error; (2) $\sigma$ used in the calculation is transformation uncertainty, while intensities over patches is label uncertainty \cite{Jax2}. These two quantities may be inherently uncorrelated in GP registration. (3) Features that surgeons are interested in, e.g., tumor margins or nearby blood vessels, may be limited to a small region. It may make more sense to estimate the regional $\rho_{\mathrm{s}}(\Omega)$ instead of using the whole image.

\begin{figure}[t]
	\centering
	\includegraphics[height=3.6cm]{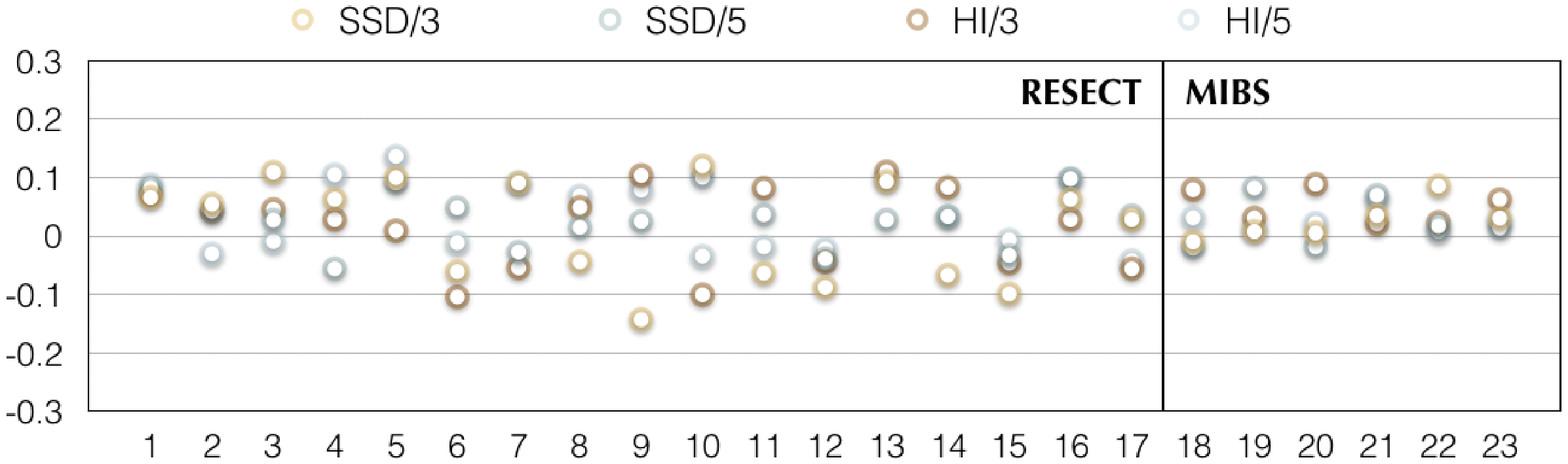}
		\vspace{-4mm}
	\caption{The estimated $\rho_{\mathrm{s}}(\Omega)$ for the patch-wise experiment. Values of $\rho_{\mathrm{s}}(\Omega)$'s are consistently low for both datasets. }
		\vspace{-3mm}
\end{figure}

\vspace{-2mm}
\section{Conclusion}
\vspace{-2mm}

``Are registration uncertainty and error monotonically associated?' is a fundamental question that has been overlooked by researchers in the medical imaging community. There has been significant progress in the development of fast and accurate methods for performing non-rigid registration. Since all of these methods are subject to some error and rarely used in the operating room, an answer to this question, which enables the use of registration uncertainty as a surrogate for assessing registration error, can increase the feasibility of non-rigid registration in interventional guidance and advance the state of image-guided therapy.

In this pilot study, we systematically investigate the monotonic association between Gaussian process registration uncertainty and error in the context of Image-guided neurosurgery. At the current stage, the low-to-moderate correlation between GP uncertainty and error indicates that it may not be feasible to apply it in practice. Nevertheless, this work opens a research area for uncertainty/error relationship analysis and may inspire more research on this topic to verify and enhance the applicability of registration uncertainty.

\end{document}